%% file: main.tex
\documentclass[]{gmaart2022}

\renewcommand{\varProcName}{Proc. 32. Workshop Computational Intelligence}
\renewcommand{\varLocation}{Berlin}
\renewcommand{\varDate}{01.-02.12.2022}

\usepackage[utf8]{inputenc} 
\usepackage[T1]{fontenc} 
 
\usepackage{amsmath,amssymb,amsfonts,mathtools,amsthm} 

\usepackage[babel,autostyle]{csquotes}

\usepackage{tabularx}
\newcolumntype{L}[1]{>{\raggedright\arraybackslash}p{#1}} 
\newcolumntype{C}[1]{>{\centering\arraybackslash}p{#1}} 
\newcolumntype{R}[1]{>{\raggedleft\arraybackslash}p{#1}} 
\usepackage{booktabs}

\usepackage{paralist}   

\usepackage[noend]{algpseudocode}

\algrenewcommand\algorithmicrequire{\textbf{Voraussetzung:}}
\algrenewcommand\algorithmicensure{\textbf{Abschlussbedingung:}}

\usepackage{listings} 
\usepackage{subcaption}  

\usepackage{epstopdf}
\usepackage{xcolor}

\usepackage{scrhack}
\usepackage{varwidth}
\usepackage{rotating} 
\usepackage[defaultlines=2,all]{nowidow}  
\usepackage{import}
\usepackage{placeins}
\usepackage{makecell}


\input{custompackages.tex}

\begin{document}


\hyphenpenalty=2000

\pagenumbering{roman}
\cleardoublepage
\setcounter{page}{1}
\pagestyle{scrheadings}
\pagenumbering{arabic}

\setnowidow[2]
\setnoclub[2]

\input{beitrag}

\addtocontents{toc}{\protect\newpage}



\end{document}

%% file: custompackages.tex
\usepackage{xspace}
\usepackage{xcolor,colortbl}
\usepackage{moresize}
\usepackage{hyperref}
\usepackage{cleveref}
\usepackage{multirow}
\usepackage{enumitem}
\usepackage{listings}

\usepackage{pifont}
\newcommand{\cmark}{\ding{51}}%
\newcommand{\xmark}{\ding{55}}%
\newcommand{\CELLBAD}{\xmark}
\newcommand{\CELLGOOD}{\cmark}

\usepackage{etoolbox}
\preto\section\glsresetall

\usepackage{acronym}
\usepackage{glossaries}

\newacro{DNN}[DNN]{Deep Neural Network}
\newacro{ML}[ML]{Machine Learning}
\newacro{GUI}[GUI]{Graphical User Interface}
\newacro{GPU}[GPU]{Graphical Processing Unit}
\newacro{REST}[REST]{Representational State Transfer}
\newacro{OPSD}[OPSD]{Open Power System Data}
\newacro{DEEPaaS}[DEEPaaS]{DEEP as a Service}

\newcommand\ie{i.\,e.\xspace}
\newcommand\eg{e.\,g.\xspace}


%% file: beitrag.tex
\renewcommand{\Title}{EasyMLServe: Easy Deployment of REST Machine Learning Services}

\renewcommand{\Authors}{
    Oliver Neumann, Marcel Schilling, Markus Reischl, Ralf Mikut
}
\renewcommand{\Affiliations}{
    Institute for Automation and Applied Informatics, \\
    Karlsruhe Institute of Technology,
    Hermann-von-Helmholtz-Platz 1, \\
    76344 Eggenstein-Leopoldshafen \\
    E-Mail: oliver.neumann@kit.edu
}

\renewcommand{\AuthorsTOC}{O.~Neumann, M.~Schilling, M.~Reischl, R.~Mikut} 
\renewcommand{\AffiliationsTOC}{Karlsruhe Institute of Technology, Institute for Automation and Applied Informatics} 

\setLanguageEnglish
							 
\setupPaper 

\begin{sloppy}

\section*{Abstract}

Various research domains use machine learning approaches because
they can solve complex tasks by learning from data.
Deploying machine learning models, however, is not trivial
and developers have to implement complete solutions
which are often installed locally and include Graphical User Interfaces (GUIs).
Distributing software to various users on-site has several problems.
Therefore, we propose a concept to deploy software in the cloud.
There are several frameworks available based on Representational State Transfer (REST)
which can be used to implement cloud-based machine learning services.
However, machine learning services for scientific users have special
requirements that state-of-the-art REST frameworks do not cover completely.
We contribute an EasyMLServe software framework to deploy machine learning
services in the cloud using REST interfaces and generic local or web-based GUIs.
Furthermore, we apply our framework on two real-world applications,
\ie, energy time-series forecasting and cell instance segmentation.
The EasyMLServe framework and the use cases are available on GitHub.

\section{Introduction}

\ac{ML} approaches are able to solve complex tasks in various domains by learning from data,
for example, instance segmentation \cite{stringer_cellpose_2021}, translation \cite{vaswani_attention_2017}, or text-to-image generation \cite{ramesh_zero-shot_2021}.
Nonetheless, users struggle to apply \ac{ML} approaches to their data
because adapting code and setting up software and hardware environments need expert knowledge \cite{gomez-de-mariscal_deepimagej_2021}.
Therefore, it is important to deploy \ac{ML} models in a way such that non-expert users
can apply these models to their data easily.

Typically, \ac{ML} models are deployed by programming \acp{GUI}
because users are familiar with it \cite{gomez-de-mariscal_deepimagej_2021, belda_datimes_2020, doty_design_2022}. 
This deployment strategy, however, has some disadvantages.
Each individual user needs powerful hardware because \ac{ML} models are computationally expensive.
The hardware device is often underutilized as users do not process data continuously.
Additionally, the hardware does not scale with user requests.
Experts have to perform the installation process on the user site
because of software dependencies, for example, GPU drivers or \ac{ML} frameworks.
Code quality suffers because of mixing model and \ac{GUI} code.
Updating software is complicated since distributed code snippets along different users.

To solve these deployment problems, we contribute a software framework (EasyMLServe) to deploy \ac{ML} approaches
as cloud-based software services using \ac{REST} \cite{fielding_rest_2000}.
Therefore, we introduce a concept of cloud-based software,
the framework architecture, and apply the framework to two real-world applications.
Additionally, the framework is available on GitHub (\url{https://github.com/KIT-IAI/EasyMLServe}), including two real-world applications.

In Section \ref{sec:relatedwork}, we introduce existing frameworks for the cloud-based deployment of \ac{ML} models.
The requirements, concept, architecture, and implementation of the contributed framework is explained in Section \ref{sec:easmlserve}.
In Section \ref{sec:results}, we evaluate the frameworks and apply EasyMLServe on two real-world use cases,
\ie, energy time-series forecasting and instance segmentation of biological images.
Afterwards, we discuss and conclude the results in Section \ref{sec:discussion} and Section \ref{sec:conclusion}.

\section{Related Work}
\label{sec:relatedwork}

There are several \ac{REST} frameworks available to deploy \ac{ML} services.
A lot of them focus on high-performance
like TorchServe \cite{bafna_torchserve} or TFX Serving \cite{gorovoy_tfxserving}.
Other frameworks offer easy-to-use config-based deployment of \ac{REST} \ac{ML} frameworks like \ac{DEEPaaS} \cite{lopez_garcia_deepaas_2019}.

TorchServe is a framework to deploy \ac{ML} services in the cloud
and it is part of the PyTorch ecosystem \cite{paszke_pytorch_2019}.
The framework is actively maintained and allows parallel requests
as well as advanced features like model performance optimization.
TorchServe offers a broad range of examples due to the large community.
However, TorchServe is restricted to the PyTorch ecosystem and excludes other \ac{ML} frameworks
like TensorFlow \cite{abadi_tensorflow_2015} or Scikit-Learn \cite{pedregosa_scikit-learn_2011}.

TFX Serving is part of the TensorFlow Extended (TFX) framework for deploying productive \ac{ML} pipelines
and is also part of the TensorFlow ecosystem.
TFX Serving is also actively maintained and supports parallel requests, advanced features, and offers a variety of examples.
However, like TorchServe, it is not independent of the \ac{ML} framework, has a complex interface and documentation,
and no \ac{GUI} support.

\ac{DEEPaaS} is an independent framework to deploy \ac{ML} services in the cloud.
It is actively maintained and it offers a simple config-based interface description. 
\ac{DEEPaaS} has support for multiple workers but it does not handle multiple GPU access.
Therefore, \ac{DEEPaaS} recommends running only one worker if there is one GPU.
Additionally, \ac{DEEPaaS} offers no examples and no \ac{GUI} support.


There are more \ac{REST} frameworks for \ac{ML} services available
but all of them are very similar to the presented frameworks and have the same problems:
They are focused on performance and, thus, are not independent of the \ac{ML} framework.
They offer many additional features, e.g., model management, which makes them complex,
and they support no generic \acp{GUI} to support fast prototyping.
With EasyMLServe, we want to offer an independent and easy-to-use
\ac{REST} framework for \ac{ML} services that additionally deploys generative
\acp{GUI} as a starting point for non-experts users.

\section{EasyMLServe}
\label{sec:easmlserve}

In this section, we introduce the EasyMLServe framework to deploy
\ac{ML} services using \ac{REST} APIs and generic \acp{GUI}.
First, we define the requirements of \ac{REST} frameworks for scientific \ac{ML} services.
Second, we describe the basic concept of \ac{REST} APIs and how the data is exchanged between the \ac{ML} service, the server, and the \ac{GUI}.
Third, we describe the most important EasyMLServe classes.
Fourth, we explain how developers can deploy their \ac{ML} approaches with the EasyMLServe framework.

\subsection{Requirements}

Developers in charge of implementing \ac{ML} services for scientific users from different domains
have requirements for \ac{REST} frameworks to deploy their \ac{ML} services.
\begin{itemize}
    \item The \ac{REST} framework should be actively maintained to ensure version compatibility, continuous improvement, and bug fixing.
    \item As developers of \ac{ML} approaches for researchers, it is necessary to use different \ac{ML} frameworks
          because not every novel \ac{ML} approach is available in all \ac{ML} frameworks.
          Therefore, a \ac{REST} framework independent of the \ac{ML} framework is required.
    \item In the research domain, prototypes of \ac{ML} approaches need to be developed fast
          in the dynamic \ac{ML} research community.
          Fast development can be achieved by offering easy-to-use and accessible frameworks.
    \item Non-expert users need a fast and easy way to use the \ac{ML} approach.
          \acp{GUI} help to fulfill that objective.
    \item Real-world \ac{ML} examples are needed to give developers a good starting point for their approaches.
          This makes the framework easier to use and reduces development time.
\end{itemize}

Other \ac{REST} frameworks for \ac{ML} services, in general, have additional requirements.
However, those requirements are not needed in our use cases.
Hence, we classify them as optional requirements:
\begin{itemize}
    \item Industrial \ac{ML} services need \ac{REST} frameworks that handle thousands or more users in parallel
          which is challenging when thinking of large high-performance computing clusters on which the service should run.
          Examples of such applications can be found at Amazon, Spotify, or Netflix.
    \item Some \ac{REST} frameworks offer more advanced features like model management
          that always come with additional code and configuration effort.
\end{itemize}

\subsection{Concept}

Based on the requirements, we propose a cloud-based service-oriented software architecture.
Each \ac{ML} approach is a software program (service) running on a remote computer (cloud).
Data between services and users are exchanged using \ac{REST}.

\ac{REST} is a design principle for distributed systems and is based on the HTTP method stack \cite{fielding_rest_2000}.
Therefore, it can be used for every network, \eg, the internet or local private networks,
and on any device, from smart meters to high-performance clusters.
Software services that are based on \ac{REST} often exchange data using JSON files
which are equivalent to a list of key-value pairs.

If we apply the \ac{REST} principle to our use case,
we have a server offering a \ac{ML} service by providing a \ac{REST} interface.
All communications between the \ac{GUI} and server are based on HTTP messages.
The \ac{REST} interface, however, has no \ac{GUI}.
Therefore, our concept considers a \ac{GUI} to communicate with the server via the \ac{REST} interface.
This \ac{GUI} can be in form of a web page or a program running on a local computer.
An overview of the data exchange between users and services is shown in Figure \ref{fig:data_transfer}.

\begin{figure}
    \centering
    \includegraphics[width=0.95\textwidth]{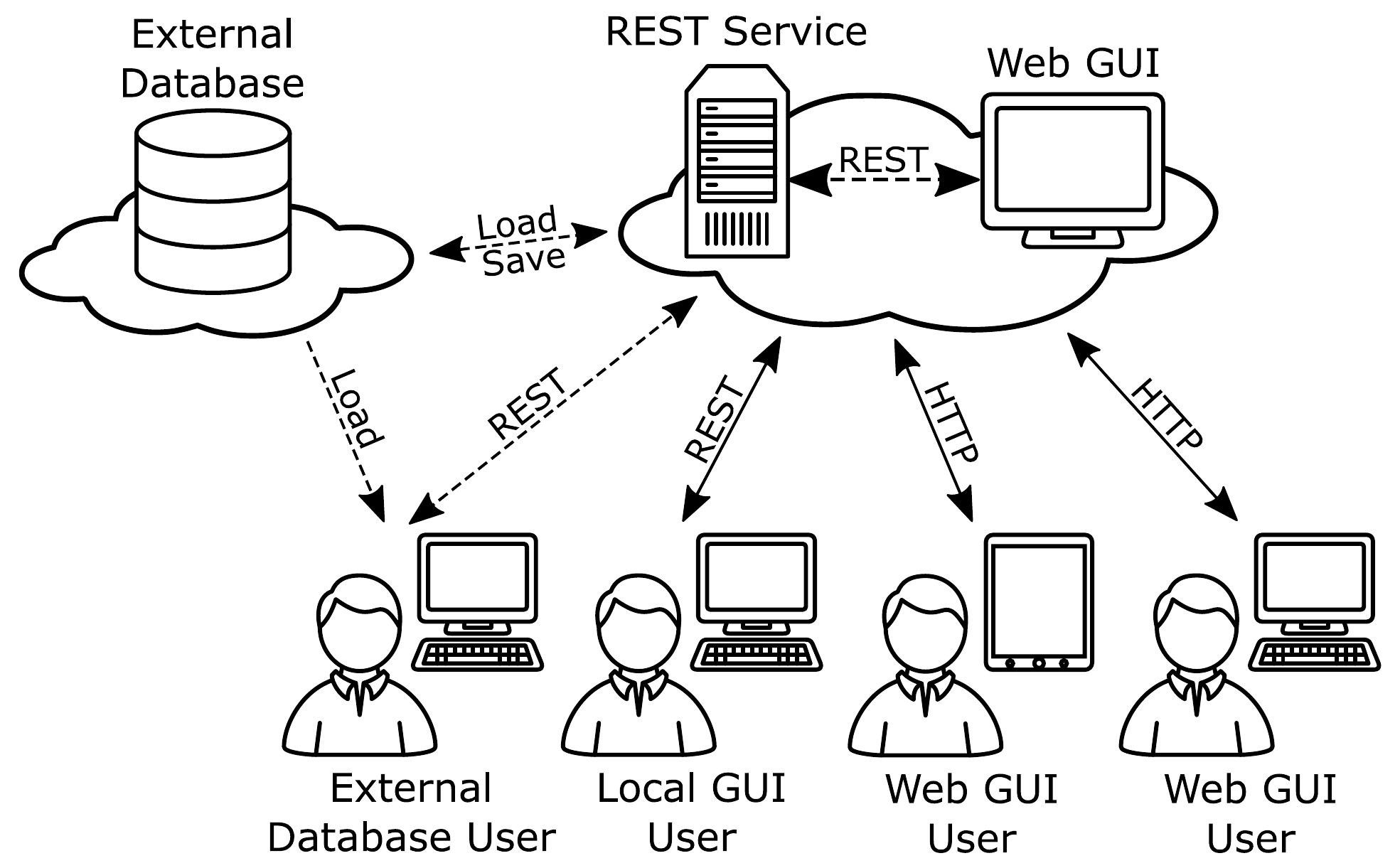}
    \caption{
        Data exchange between users and \ac{ML} services.
        All data exchange between a user and a service is based on JSON objects.
        \acp{GUI} are deployed remotely as a web page or locally on a computer.
        Without \acp{GUI}, smart meters, or other low-end devices, can directly
        use the \ac{ML} service using the \ac{REST} interface.
        Additionally, data exchange between third-party actors is possible
        by using, for example, additional databases.
    }
    \label{fig:data_transfer}
\end{figure}

For the EasyMLServe framework, we decided that all requests and responses
of the \ac{REST} interface are encoded as JSON objects.
It would be possible to exchange data directly as files but
that increases the framework complexity.
Therefore, all data from and to the \ac{ML} services are encoded as JSON objects
which makes it easier for developers and reduces framework complexity.

Depending on the \ac{GUI} type, there are two ways of data exchange between users and the \ac{GUI}.
First, for web-based \acp{GUI}, we propose to communicate with the \ac{GUI} via HTTP
but not using a \ac{REST} interface. Instead, the \ac{GUI} is deployed using a 
web server that displays a web page to use the \ac{ML} service.
Second, for local \acp{GUI}, our concept suggests interacting directly with the \ac{GUI} without using HTTP.
In both cases, however, the \ac{GUI} has to communicate with the \ac{ML} service
using the \ac{REST} interface.

Data can also be exchanged with additional actors using any kind of communication protocol.
A simple example would be to load and store data from a database instead of
uploading data via HTTP which can accelerate processing.

Regarding hardware constraints, the server needs to be deployed on powerful hardware
depending on the \ac{ML} task. For instance segmentation tasks, a \ac{GPU} is recommended
but for time-series forecasting using Linear Regression, a CPU is sufficient.
The \ac{GUI}, however, can be deployed on low-end hardware.
Additionally, if a \ac{GUI} is not needed, low-end devices can directly communicate with the \ac{REST} interface.
Smart meters, for example, can forecast energy time-series data or
microscopes can process images by using high-performance computing clusters.

\subsection{Architecture}

The EasyMLServe framework consists of three major classes,
\ie, \textit{EasyMLService}, \textit{EasyMLServer}, and \textit{EasyMLUI}.
The actual \ac{ML} service is represented by the \textit{EasyMLService} class.
The \textit{EasyMLServer} deploys the \ac{REST} interface.
\textit{EasyMLUI} is the base class for all other UI classes.
The relation between these three classes is shown in Figure \ref{fig:uml_classes}.

\begin{figure}
    \centering
    \includegraphics[width=\textwidth]{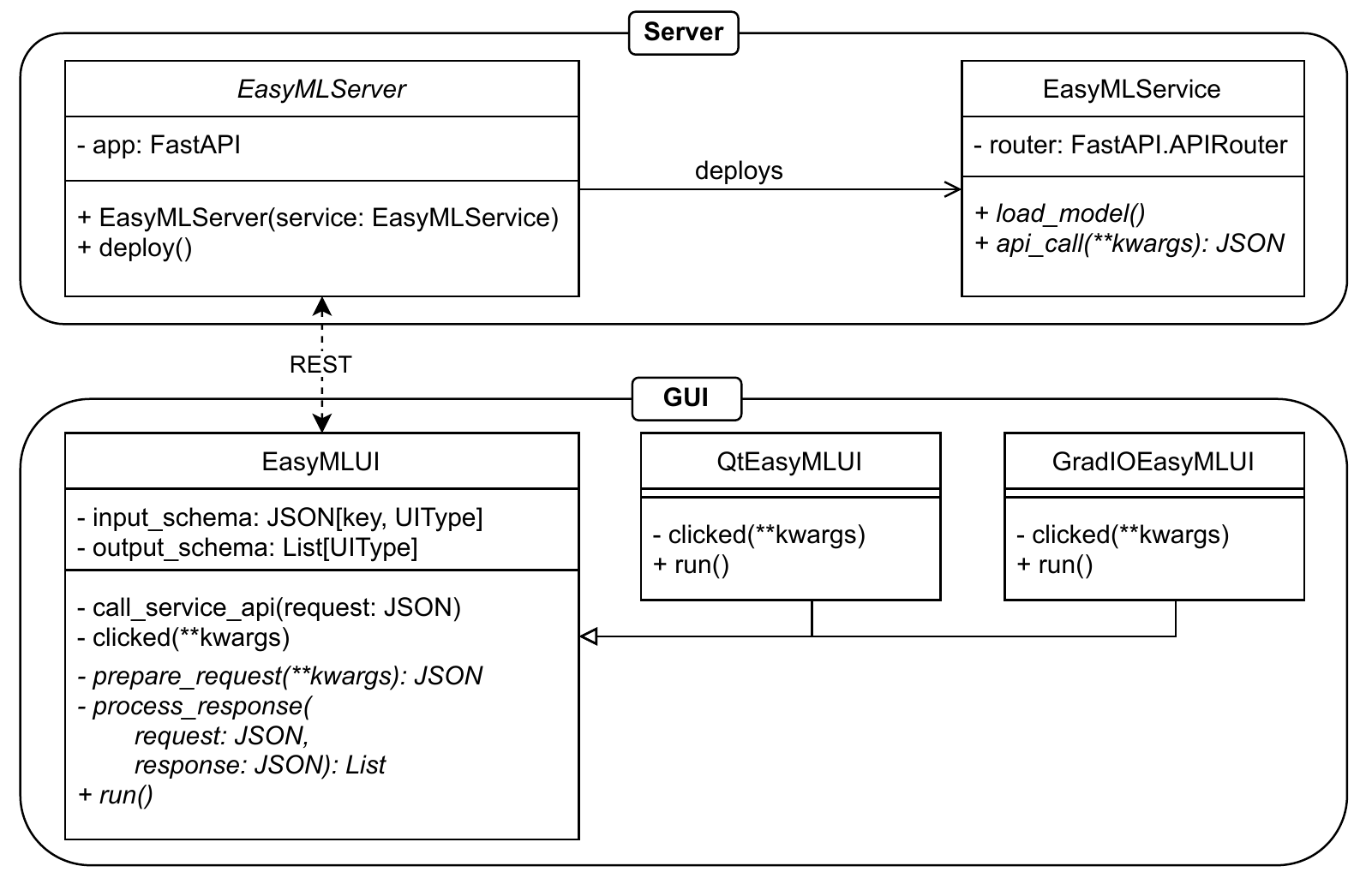}
    \caption{
        The basic architecture of EasyMLServe with the most important attributes and methods.
        The top block (Server) shows the relation between \textit{EasyMLServer} and \textit{EasyMLService}.
        The bottom part (GUI) shows the \ac{GUI} classes and their relation.
        Both parts are separated such that an \textit{EasyMLService} can be deployed
        by the server without using an \textit{EasyMLUI} \ac{GUI}.
    }
    \label{fig:uml_classes}
\end{figure}

\textit{EasyMLService} is responsible for loading the model
and processing JSON request of the \ac{REST} interface.
It returns JSON objects as a result of the \ac{REST} interface request.

\textit{EasyMLServer} provide the actual \ac{REST} interface and passes all \ac{REST} requests to the \textit{EasyMLService}.
Therefore, the \textit{EasyMLServer} has to deploy a web server that handles HTTP messages
which is done by the Uvicorn framework \cite{christie_uvicorn}.

\textit{EasyMLUI} is the base class for all available generic \acp{GUI} of the EasyMLServe framework.
It handles the exchange of data between the user and the actual \ac{REST} \ac{ML} service.
Therefore, \textit{EasyMLUI} takes user input, prepares a \ac{REST} request, and gets a \ac{REST} response
by passing the \ac{REST} request to the \ac{REST} \ac{ML} service via HTTP.
Currently, we support two frameworks for \acp{GUI}, \ie, PyQt \cite{ottosson_pyqt} and Gradio \cite{abid_gradio}.
Both \acp{GUI} are available using the child classes \textit{QtEasyMLUI} and \textit{GradioEasyMLUI}.

\subsection{Implementation}

Developers that use EasyMLServe need to implement two classes:
A service class with \textit{EasyMLService} as parent class
and a \ac{GUI} class with \textit{QtEasyMLUI} or \textit{GradioEasyMLUI} as parent class.
All relevant classes and their methods are shown in Figure \ref{fig:uml_classes}.

EasyMLService consists of two methods developers need to implement, \ie, \texttt{load\_model} and \texttt{api\_call}.
The \texttt{load\_model} method is called after the \textit{EasyMLService} is initialized
and allows the user to load the model.
The \texttt{api\_call} method is called when a \ac{REST} request is received with the actual JSON object.
In code, developers have to implement a \textit{EasyMLService} class like:
\begin{lstlisting}[basicstyle=\fontsize{8}{9}\selectfont\ttfamily]
class MyMLService(EasyMLService):

    def load_model(self):
        # load and prepare model
        pass

    def process(self, request):
        response = {...}  # prepare REST response 
        return response
\end{lstlisting}

\textit{EasyMLUI} is the base class for all other \ac{GUI} classes.
Every \ac{GUI} class has to implement two methods, \ie, \texttt{}{prepare\_request} and \texttt{process\_response}.
The \texttt{prepare\_request} method gets all user inputs defined in the input scheme and returns the \ac{REST} request JSON encoded.
The \texttt{process\_response} method prepares and returns results that should be displayed to the user.
In code, developers have to implement a \textit{EasyMLUI} class like:
\begin{lstlisting}[basicstyle=\fontsize{8}{9}\selectfont\ttfamily]
class MyMLServiceUI(EasyMLUI):

    def prepare_request(self, some_ui_input):
        request = {...}  # prepare REST request
        return request

    def process_response(self, request, response):
        results = ...  # prepare results (e.g. plots)
        return results
\end{lstlisting}

To initialize \textit{EasyMLUI} classes users have to define an input and output scheme.
Input schemes describe the data users pass to the \ac{GUI}, \eg, a CSV file.
Output schemes determine what the \ac{GUI} is presenting to the user,
\eg, segmentation results or evaluation files.
We need to define an input scheme as data can be passed in several ways, for example,
text for translation tasks can be passed via text files or by typing text into text boxes.
The output scheme is needed because displaying the response is often not sufficient, for example,
when displaying segmentation results users may also be interested in the number of cells or mean cell size.

To define the input and output scheme, we implemented a set of \textit{UITypes} which produce suitable \ac{GUI} elements.
These \textit{UITypes} can be:
\textit{Text}, \textit{TextLong}, \textit{Number}, \textit{Range}, \textit{SingleChoice}, \textit{MultipleChoice},
\textit{File}, \textit{ImageFile}, \textit{CSVFile}, \textit{TimeSeriesCSVFile}, or \textit{Plot}.

After implementing an \textit{EasyMLService} and \textit{EasyMLUI} the resulting \ac{ML} service
and \ac{GUI} can be deployed by calling the \texttt{run} method.
In code, starting the server and service looks like:
\begin{lstlisting}[basicstyle=\fontsize{8}{9}\selectfont\ttfamily]
# server.py
service = MyMLService()
server = EasyMLServer(service)
server.run()
\end{lstlisting}
\begin{lstlisting}[basicstyle=\fontsize{8}{9}\selectfont\ttfamily]
# ui.py
input_schema = {'some_ui_input': UIType()}
output_schema = [Plot()]
app = MyMLServiceUI(name='Example Service',
                    input_schema=input_schema,
                    output_schema=output_schema)
app.run()
\end{lstlisting}

\section{Results}
\label{sec:results}

In the following, we evaluate the presented \ac{REST} frameworks
TorchServe \cite{bafna_torchserve}, TFX Serve \cite{gorovoy_tfxserving}, \ac{DEEPaaS} \cite{lopez_garcia_deepaas_2019}, and EasyMLServe
based on the previously defined requirements for \ac{REST} frameworks
in scientific domains.
Afterwards, we apply the EasyMLServe framework on two real-world applications,
\ie, energy time-series forecasting and cell instance segmentation,
to demonstrate the benefits of the proposed framework.

\subsection{Evaluation}

EasyMLServe is a \ac{REST} framework for \ac{ML} services
that is focused on deployment of \ac{ML} approaches for the research community.
Therefore, we define specific requirements which need to be solved.

Evaluating existing \ac{REST} frameworks with these requirements,
we see that \ac{REST} frameworks, which are focused on performance,
are restricted to one \ac{ML} framework, \eg, TorchServe or TFX Serve.
Other \ac{ML} frameworks like \ac{DEEPaaS}, however, are independent
of the \ac{ML} framework but offer no generative \ac{GUI} support.

Our EasyMLServe framework fulfills all requirements
and closes the gap of actively maintained, \ac{ML} framework independent,
easy-to-use, and generative \ac{GUI} supported \ac{REST} frameworks.
A comparison of the \ac{REST} frameworks based on all requirements,
including the optional ones, is shown in Table \ref{tab:framework_comparison}.

\begin{table}
    \centering
    \caption{
        \ac{REST} frameworks for \ac{ML} services evaluated on the necessary and optional requirements.
        Regarding necessary requirements, \ac{REST} frameworks need to be actively maintained,
        independent of the \ac{ML} framework, easy accessible (not complex), supports generative \acp{GUI},
        and real-world examples.
        Regarding optional requirements, \ac{REST} frameworks need to handle parallel requests and advanced features (e.g. model management).
    }
    \resizebox{\textwidth}{!}{
        \begin{tabular}{l
                        >{\centering}p{2cm}
                        >{\centering}p{2cm}
                        >{\centering}p{2cm}
                        >{\centering\arraybackslash}p{2cm}}
            \toprule
            \textbf{Requirements} & \multicolumn{4}{c}{\textbf{REST Frameworks for ML}} \\
             & TorchServe & TFX Serving & \ac{DEEPaaS} & EasyMLServe \\
            \hline
            Maintained & \CELLGOOD & \CELLGOOD & \CELLGOOD & \CELLGOOD \\ 
            Independent & \CELLBAD & \CELLBAD & \CELLGOOD & \CELLGOOD \\
            Accessible & \CELLBAD & \CELLBAD & \CELLGOOD & \CELLGOOD \\
            GUI Support & \CELLBAD & \CELLBAD & \CELLBAD & \CELLGOOD \\
            Examples & \CELLGOOD & \CELLGOOD & \CELLBAD & \CELLGOOD \\
            \hline
            Parallel & \CELLGOOD & \CELLGOOD & (\CELLGOOD) & (\CELLGOOD) \\
            Advanced Features & \CELLGOOD & \CELLGOOD & \CELLBAD & \CELLBAD \\
        \end{tabular}
    }
    \label{tab:framework_comparison}
\end{table}

Additionally, to demonstrate the capabilities of EasyMLServe, we implemented two real-world use cases.
First, electrical load forecasting for Germany which is representative for several time-series \ac{ML} problems.
Second, biological cell instance segmentation which is a common \ac{ML} task for image processing.

\subsection{Time-Series Forecasting}

In the first use case, we forecast the electrical load for Germany one day ahead.
Hourly electrical load data is used from the \ac{OPSD} dataset \cite{wiese_opsd_2019}.
We used the years 2015 to 2018 for training.
Regarding the models, we consider a Linear Regression, Support Vector Machine, and Random Forest model.
All models are implemented using Scikit-Learn \cite{pedregosa_scikit-learn_2011} with default parameter settings.

Our \ac{ML} service expects a list of model names to use for the forecast,
a list of time steps, and the corresponding energy values also as a list.
After applying the selected models on the energy time-series,
the \ac{ML} service returns a forecast for each selected model in a list.
Each forecast contains the corresponding model name, a list of time steps,
and the corresponding energy values.

Regarding the \ac{GUI} input scheme, we choose the MultipleChoice \ac{GUI} element to define the model names.
For the time steps and energy values, we use the CSVFile \ac{GUI} element.
This CSV file needs to be parsed to finally return the model names, time steps, and energy values as one JSON request.
The parsing is done in the \texttt{prepare\_request} method.

For the output scheme, we return two plots and a CSV file.
For the two plots, we use the Plot \ac{GUI} element
to visualize the forecast and forecast error.
Regarding the CSV file, a File \ac{GUI} element
which contains the actual one-day ahead forecasts is used.

Both supported \acp{GUI} can be deployed with EasyMLServe.
Developers are able to switch easily between the \acp{GUI} by changing the parent class.
The resulting \acp{GUI} are shown in Figure \ref{fig:time_series_ui}.

\begin{figure}
     \centering
     \begin{subfigure}[b]{0.99\textwidth}
         \centering
         \includegraphics[width=0.98\textwidth]{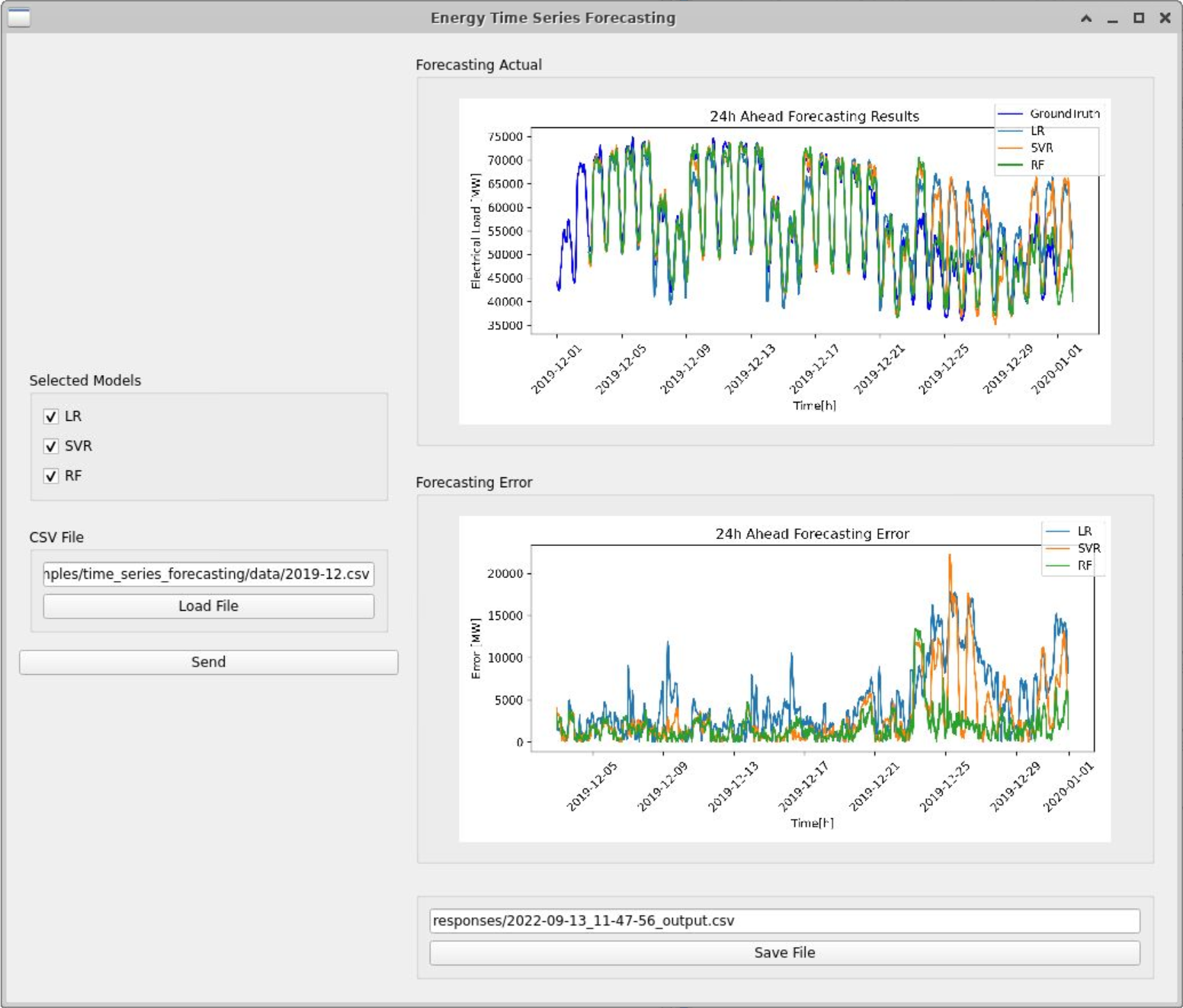}
         \caption{Qt UI}
         \label{fig:time_series_qt}
     \end{subfigure}
     \hfill
     \begin{subfigure}[b]{0.99\textwidth}
         \centering
         \includegraphics[width=0.98\textwidth]{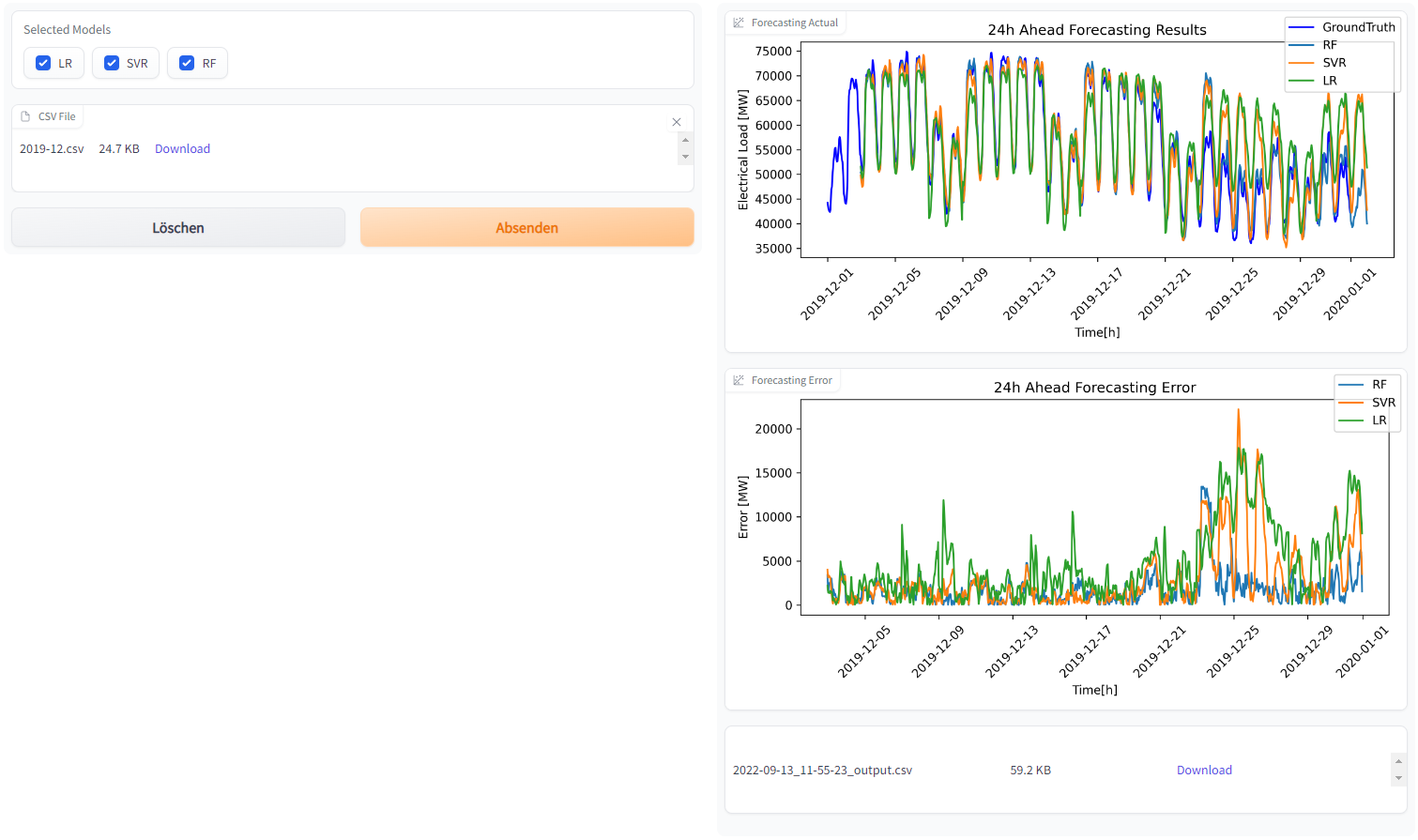}
         \caption{Gradio UI}
         \label{fig:time_series_gradio}
     \end{subfigure}
    \caption{
        Qt (a) and Gradio (b) based \acp{GUI} for the energy time-series forecasting use case.
        Qt is a locally deployed \ac{GUI}. Gradio is a web-based deployed \ac{GUI}.
    }
    \label{fig:time_series_ui}
\end{figure}

\subsection{Image Segmentation}

The second real-world use case focuses on machine learning for images.
We use microscopic images from the LIVECell dataset \cite{edlund_livecelllarge-scale_2021}
and train a UNet \cite{ronneberger_u-net_2015} utilizing the KaIDA framework \cite{schilling_kaida_2022}.

The \ac{ML} service expects a Base64 encoded image containing
the image encoding, data type, and shape.
After segmenting the cells, our service returns an image of detected instances Base64 encoded.
Note, the resulting \ac{ML} service response has the same structure as the input request.

Regarding the \ac{GUI} input scheme, we only use a File \ac{GUI} element to select the image file.
After loading the image, the image is encoded as Base64 and the \ac{REST} request is created as JSON.

For the output scheme, we want to display the instances image, number of cells,
mean cell size, and the \ac{ML} service response as a JSON file.
The instance overlay image is displayed by using the Plot \ac{GUI} element of EasyMLServe.
The number of cells and mean cell size are visualized using the Number \ac{GUI} element.
The response is displayed as a File \ac{GUI} element where a user can download it.

The \ac{GUI} generation framework of EasyMLServe supports
Qt, as a local \ac{GUI}, and Gradio, as a web-based \ac{GUI}.
Developers can switch between the two \ac{GUI} frameworks
by changing the parent class of the implemented \textit{EasyMLUI} class.
Both \acp{GUI} can be seen in Figure \ref{fig:segmentation_ui}.

\begin{figure}
     \centering
     \begin{subfigure}[b]{0.99\textwidth}
         \centering
         \includegraphics[width=0.98\textwidth]{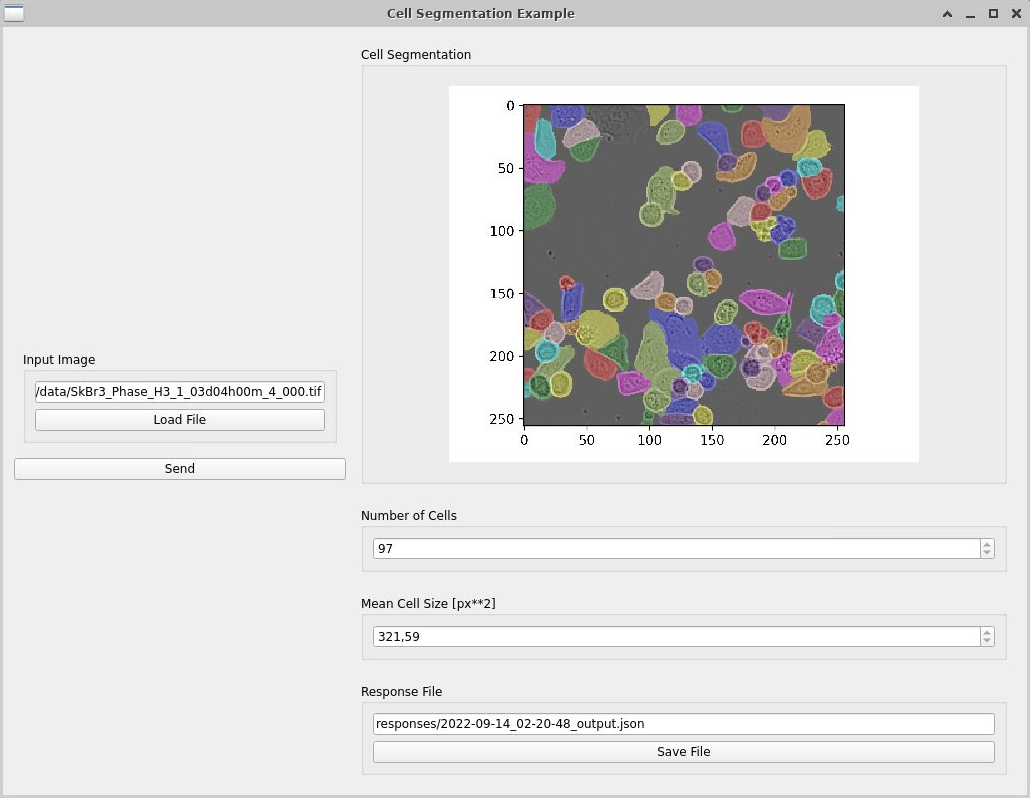}
         \caption{Qt UI}
         \label{fig:segmentation_qt}
     \end{subfigure}
     \hfill
     \begin{subfigure}[b]{0.99\textwidth}
         \centering
         \includegraphics[width=0.98\textwidth]{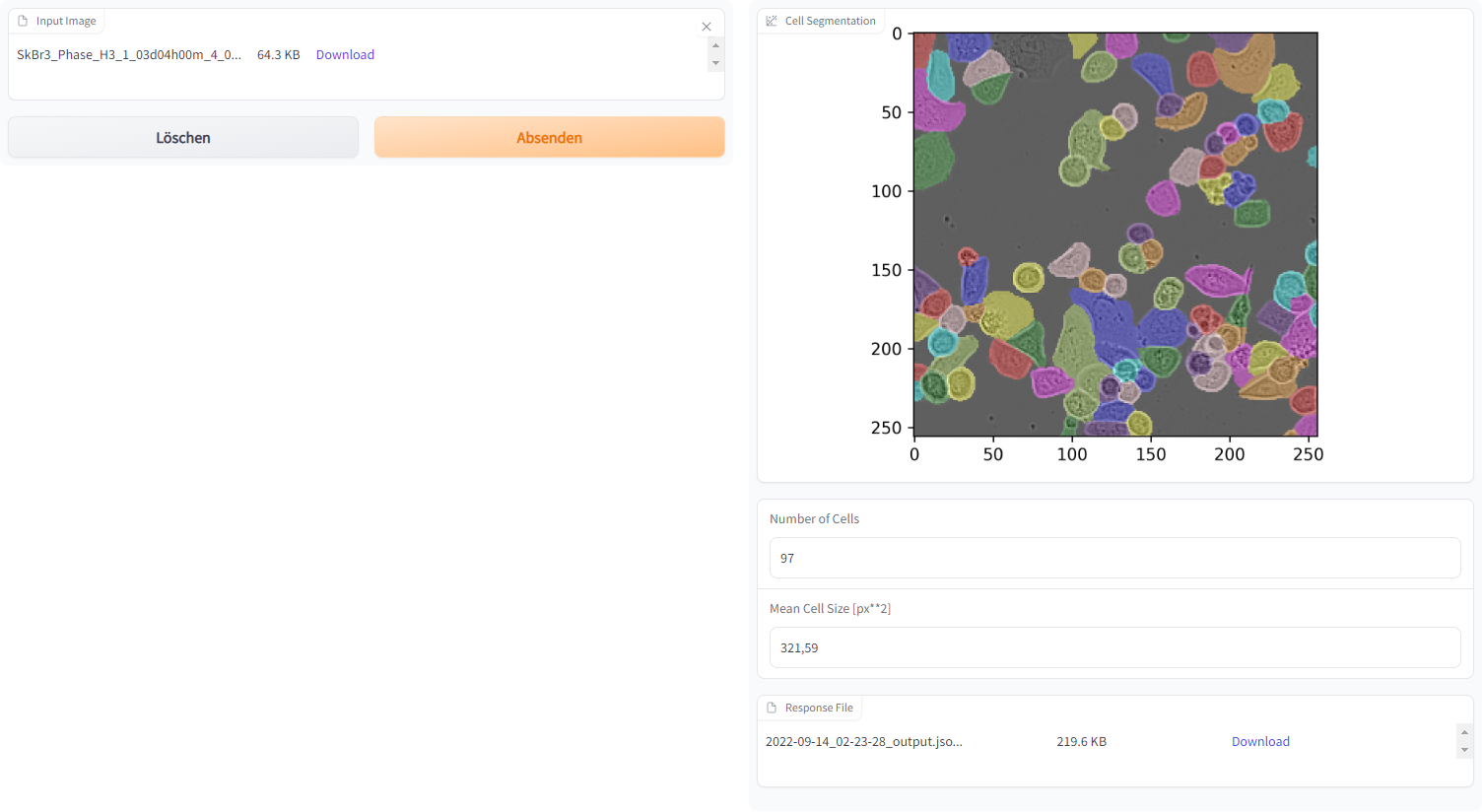}
         \caption{Gradio UI}
         \label{fig:segmentation_gradio}
     \end{subfigure}
    \caption{
        Qt (a) and Gradio (b) based \acp{GUI} for the cell segmentation use case.
        Qt is a locally deployed \ac{GUI}. Gradio is a web-based deployed \ac{GUI}.
    }
    \label{fig:segmentation_ui}
\end{figure}

\section{Discussion}
\label{sec:discussion}

EasyMLServe is a framework to easily deploy \ac{ML} services using \ac{REST}.
To reduce complexity and make the framework as slim as possible we focus on
the deployment part and leave out the training which is done by the developers in any environment.
This has the disadvantage that users need help of experts in case they want to change the running \ac{ML} service.
Some frameworks offer training routines that allow users to retrain the models.
However, we think also retraining needs the supervision of experts.
Especially in the research context where it is important that results are reliable.

The easy deployment of EasyMLServe includes generic \acp{GUI}.
These \acp{GUI} support prototyping and fast deployment.
The complexity of such \acp{GUI}, however, is limited.
It can not cover all possible \acp{GUI} without losing its accessibility.
Therefore, developers still have to develop \acp{GUI} on their own
if the limited complexity of the generic \acp{GUI} is not enough.

In EasyMLServe, \ac{ML} services exchange data using JSON objects
which is a common way for \ac{REST} services.
It is also possible to directly upload files using multipart/form-data.
This would avoid encoding and decoding files
and thus reduces processing time and package size.
We restricted the EasyMLServe \ac{REST} interface to JSON objects
because it made the framework and communication for the \ac{GUI} easier.
Furthermore, processing times of \ac{ML} services are mostly restricted to
the \ac{ML} approach itself which is usually the most computational expensive part,
\eg, instance segmentation using Deep Neural Networks running on a GPU.

For the implementation of EasyMLServe, we use Python as the programming language.
Currently, the most common \ac{ML} frameworks are written in Python.
Therefore, all recent \ac{ML} approaches are available in Python.
However, \ac{ML} approaches that are not written in Python
can not easily be integrated into the presented framework.

Finally, EasyMLServe is a novel framework and we just started with a first version.
There are bugs that need to be found and fixed
as well as features which are currently missing.
Nonetheless, bugs will appear and feature
requirements will occur when developers apply this framework
which belongs to a normal life cycle of software frameworks.

\section{Conclusion}
\label{sec:conclusion}

Scientific users have special requirements on the deployment of \ac{ML} approaches.
Deploying software solutions on-site has several disadvantages.
Therefore, we propose a cloud-based solution that is based on \ac{REST}
and define requirements of \ac{REST} frameworks for scientific usage.
These requirements are evaluated on the presented \ac{REST} frameworks.
Existing frameworks do not cover all necessary requirements completely
and thus we contribute EasyMLServe, a \ac{REST} framework
for easy deployment of \ac{ML} services in the cloud.
Additionally, our presented framework offers generic \acp{GUI} for fast and easy prototyping.

EasyMLServe is a fast solution for \ac{ML} developers to implement \ac{ML} services
in the cloud. It is actively maintained, independent of the \ac{ML} framework, easy-to-use,
supports generic local or web-based \acp{GUI}, and offers
real-world applications as a starting point for developers.

To further improve EasyMLServe, we propose to deploy existing solutions with the EasyMLServe framework,
for example, pyWATTS pipelines \cite{heidrich_pywatts_2021}.
EasyMLServe is a novel framework which is under development.
In future work, we aim to improve the EasyMLServe framework 
by fixing bugs and add additional features to enhance the user experience.
This includes more \ac{GUI} elements which need to be supported by the \ac{GUI} generator
as well as more complex \acp{GUI}.

\section*{Acknowledgments}

This project is funded by the Helmholtz Association’s Initiative and Networking Fund through Helmholtz AI, the Helmholtz Association under the Programs “Energy System Design”(ESD)  and „Natural, Artificial and Cognitive Information Processing“ (NACIP), and the German Research Foundation (DFG) under Germany’s Excellence Strategy – EXC number 2064/1 – Project number 390727645.

\end{sloppy}
